\documentclass[sigconf]{acmart}
\AtBeginDocument{%
	}

\setcopyright{acmlicensed}
\copyrightyear{2025}
\acmYear{2025}
\setcopyright{acmlicensed}\acmConference[WWW Companion '25]{Companion Proceedings of the ACM Web Conference 2025}{April 28-May 2, 2025}{Sydney, NSW, Australia}
\acmBooktitle{Companion Proceedings of the ACM Web Conference 2025 (WWW Companion '25), April 28-May 2, 2025, Sydney, NSW, Australia}
\acmDOI{10.1145/3701716.3715299}
\acmISBN{979-8-4007-1331-6/25/04}

\usepackage{pifont}
\usepackage{multirow}
\usepackage{bigstrut}
\usepackage{titlesec}

\newcommand{\cmark}{\ding{51}}
\newcommand{\xmark}{\ding{55}}

\titlespacing*{\section}{0pt}{4pt}{5pt}
\titlespacing*{\subsection}{0pt}{3pt}{3pt}
\titlespacing*{\subsubsection}{0pt}{3pt}{3pt}

\begin{document}
	\title{MemEngine: A Unified and Modular Library for Developing Advanced Memory of LLM-based Agents}

	\author{Zeyu Zhang}
	\affiliation{%
		\institution{Renmin University of China}
		\city{Beijing}
		\country{China}
	}
	\email{zeyuzhang@ruc.edu.cn}
	
	\author{Quanyu Dai}
	\affiliation{%
		\institution{Huawei Noah’s Ark Lab}
		\city{Shenzhen}
		\country{China}
	}
	\email{daiquanyu@huawei.com}
	
	\author{Xu Chen$^{\dagger}$}
	\affiliation{%
		\institution{Renmin University of China}
		\city{Beijing}
		\country{China}
	}
	\email{xu.chen@ruc.edu.cn}
	
	\author{Rui Li}
	\affiliation{%
		\institution{Renmin University of China}
		\city{Beijing}
		\country{China}
	}
	\email{lirui121200@ruc.edu.cn}

	\author{Zhongyang Li}
	\affiliation{%
		\institution{Huawei Technologies Ltd.}
		\city{Beijing}
		\country{China}
	}
	\email{lizhongyang6@huawei.com}
	
	\author{Zhenhua Dong}
	\affiliation{%
		\institution{Huawei Noah’s Ark Lab}
		\city{Shenzhen}
		\country{China}
	}
	\email{dongzhenhua@huawei.com}
	
	\thanks{$\dagger$ Corresponding author.}
	
	\begin{abstract}
		Recently, large language model based (LLM-based) agents have been widely applied across various fields. As a critical part, their memory capabilities have captured significant interest from both industrial and academic communities. Despite the proposal of many advanced memory models in recent research, however, there remains a lack of unified implementations under a general framework. To address this issue, we develop a unified and modular library for developing advanced memory models of LLM-based agents, called MemEngine. Based on our framework, we implement abundant memory models from recent research works.
		Additionally, our library facilitates convenient and extensible memory development, and offers user-friendly and pluggable memory usage.
		For benefiting our community, we have made our project publicly available at \url{https://github.com/nuster1128/MemEngine}.
	\end{abstract}

	\ccsdesc[500]{Information systems~Users and interactive retrieval}
	
	\ccsdesc[500]{Software and its engineering~Software libraries and repositories}

	\keywords{Large Language Model, Autonomous Agent, Memory Mechanism, Information Retrieval, Library Resource}

	\maketitle
	
	\section{Introduction}
	With the rapid development of large language models (LLMs), LLM-based agents have been widely applied across various fields, due to their capabilities to perform complex tasks and fulfill different roles~\cite{wang2024survey}.
	Among various internal modules, memory is one of the most critical components for agents, as it determines how they store historical data, reflect on existing knowledge, and recall useful information to support decision-making~\cite{zhang2024survey}.
	Specifically, in complex tasks, memory enables the recording of critical information in the past agent-environment interactions, and provides task-related experiences from previous trajectories. In role-playing and social simulations, it highlights the characteristic and personality of each role, allowing for distinctiveness among different roles.
	
	Although some recent works have proposed various memory models for LLM-based agents, they are implemented under different pipelines and lack a unified framework. This inconsistency presents challenges for developers to attempt different models in their experiments.
	Moreover, many basic functions (such as retrieval and summarization) are duplicated across different models, and researchers often need to implement them repeatedly when developing new models.
	Besides, many academic models are tightly integrated with agents in a non-pluggable manner, making them difficult to apply across different agents.
	
	In order to address the above problems, we develop a unified and modular library named \textbf{MemEngine}, which facilitates the development of advanced memory models for LLM-based agents. The primary features of our library are summarized as follows:
	
	\textbf{Unified and Modular Memory Framework.}
	We propose a unified memory framework composed of three hierarchical levels to organize and implement existing research models under a general structure.
	The lowest level comprises memory functions, implementing basic functions (\textit{e.g.,} retrieval) as foundational supports for different memory operations.
	The intermediate level encompasses memory operations, constituting basic operations (\textit{e.g.,} memory recall) to construct different memory models.
	The highest level involves memory models, implementing various existing research models (\textit{e.g.,} MemoryBank~\cite{zhong2024memorybank}) that can be conveniently applied in different agents.
	All these three levels are modularized inside our framework, where higher-level modules can reuse lower-level modules, thereby improving efficiency and consistency in implementation.
	Besides, we provide a configuration module for easy modification of hyper-parameters and prompts at different levels.
	We also implement a utility module to conveniently save and demonstrate memory contents.
	
	\textbf{Abundant Memory Implementation.}
	Based on our unified and modular framework, we implement a wide range of memory models from recent research works, many of which are widely applied in diverse applications.
	All of these models can be easily switched and tested under our framework, with different configurations of hyper-parameters and prompts that can be adjusted for better application across various agents and tasks.
	
	\textbf{Convenient and Extensible Memory Development.}
	Based on our modular memory operations and memory functions, researchers can conveniently develop their own advanced memory models.
	They can also extend existing operations and functions to develop their own ones.
	To better support researchers' development, we provide detailed instructions and examples in our document to guide the customization.
	
	\textbf{Pluggable and User-friendly Memory Usage.}
	Our library offers multiple deployment options to empower LLM-based agent with powerful memory capabilities.
	Besides, we provide various memory usage modes, including default, configurable, and automatic modes.
	Moreover, our memory modules are pluggable and can be easily utilized across different agent frameworks.
	Our library is also compatible with some prominent frameworks of LLM-based agents, such as AutoGPT.
	These features collectively contribute to making our library more user-friendly.
	
	In summary, MemEngine is the first library that implements a wide variety of memory models from research works under a unified and modular framework, facilitating both convenient development and ease of use.
	To further benefit the community, we have made our project publicly available at Github repository\footnote{\url{https://github.com/nuster1128/MemEngine}}. Additionally, we have also organized a comprehensive documentation\footnote{\url{https://memengine.readthedocs.io/en/latest/}} for both application and development purposes.
	
	\section{Comparison with Relevant Libraries}
	Several existing libraries can also empower memory capabilities for LLM-based agents, including (1) memory modules integrated into agent libraries, and (2) independent memory libraries.
	Specifically, AutoGen\footnote{\url{https://github.com/microsoft/autogen}}, MetaGPT~\cite{hong2023metagpt}, CAMEL~\cite{li2023camel}, AgentScope~\cite{gao2024agentscope}, LangChain\footnote{\url{https://github.com/langchain-ai/langchain}}, AgentLite~\cite{liu2024agentlite}, CrewAI~\footnote{\url{https://github.com/crewAIInc/crewAI}}, AutoGPT~\footnote{\url{https://github.com/Significant-Gravitas/AutoGPT}}, and AgentVerse~\cite{chen2023agentverse} are prominent open-source libraries for building LLM-based AI agent systems.
	Besides, Memary\footnote{\url{https://github.com/kingjulio8238/Memary}} is an open-source library to empower AI agents with memory for continuous improvement. 
	Cognee\footnote{\url{https://github.com/topoteretes/cognee}} provides a scalable and modular pipeline to interconnect and retrieve previous information for AI applications.
	Mem0\footnote{\url{https://github.com/mem0ai/mem0}} offers an artificial memory layer for LLM-based agents and assistants to make them personalized.
	Agentmemory\footnote{\url{https://github.com/ai16z/agentmemory}} implements easy-to-use memory for LLM-based agents with document search and more.
	
	We present a comprehensive comparison between MemEngine and relevant libraries in Table~\ref{tab:comparison}.
	While most of these libraries offer plug-and-play memory components to store and recall information for LLM-based agents, few of them implement advanced memory operations like reflection and optimization.
	Moreover, compared with other libraries, as a major contribution, MemEngine implements comprehensive research models under a unified framework, along with providing modular operations and functions that assist researchers in customizing advanced memory models.
	
	\begin{table*}[htbp]
		\centering
		\caption{Comparison with relevant open-source libraries. We focus on both memory modules integrated in prominent agent libraries, and independent memory libraries. Besides the libraries mentioned below, AutoGPT and AgentVerse do not specify their memory support in their library. AutoGen supports MemGPT, Mem0, and Zep as extensions.}
		\vspace{-0.25cm}
		\resizebox{\linewidth}{!}{
			\begin{tabular}{cccccccc}
			\hline
			\hline
			\multirow{2}[4]{*}{\textbf{Memory Features}} & \multicolumn{7}{c}{\textbf{Memory Integrated in Agent Libraries}} \bigstrut\\
			\cline{2-8}\multicolumn{1}{c}{} & \textbf{AutoGen} & \textbf{MetaGPT} & \textbf{CAMEL} & \textbf{AgentScope} & \textbf{Langchain} & \textbf{AgentLite} & \textbf{CrewAI} \bigstrut\\
			\hline
			Plug-and-play Integration & \cmark & \xmark & \cmark & \cmark & \cmark & \cmark & \cmark \bigstrut[t]\\
			Basic Read and Write Support & \cmark & \cmark & \cmark & \cmark & \cmark & \cmark & \cmark \\
			Reflection and Optimization Support & \cmark & \xmark & \xmark & \xmark & \xmark & \xmark & \xmark \\
			Comprehensive Default Models & \xmark & \xmark & \xmark & \xmark & \xmark & \xmark & \xmark \\
			Advanced Model Customization & \xmark & \xmark & \cmark & \xmark & \xmark & \xmark & \xmark \bigstrut[b]\\
			\hline
			\multirow{2}[4]{*}{\textbf{Memory Features}} & \multicolumn{7}{c}{\textbf{Independent Memory Libraries}} \bigstrut\\
			\cline{2-8}\multicolumn{1}{c}{} & \textbf{Memary} & \textbf{Cognee} & \textbf{Mem0} & \textbf{Agentmemory} & \textbf{MemoryScope} & \textbf{Zep} & \textbf{MemEngine (Ours)} \bigstrut\\
			\hline
			Plug-and-play Integration & \cmark & \cmark & \cmark & \cmark & \cmark & \cmark & \cmark \bigstrut[t]\\
			Basic Read and Write Support & \cmark & \cmark & \cmark & \cmark & \cmark & \cmark & \cmark \\
			Reflection and Optimization Support & \xmark & \xmark & \xmark & \xmark & \cmark & \xmark & \cmark \\
			Comprehensive Default Models & \xmark & \xmark & \xmark & \xmark & \xmark & \xmark & \cmark \\
			Advanced Model Customization & \xmark & \xmark & \xmark & \xmark & \cmark & \xmark & \cmark \bigstrut[b]\\
			\hline
			\hline
			\end{tabular}
		}
		\vspace{-0.35cm}
		\label{tab:comparison}%
	\end{table*}%
	
	\section{MemEngine Library}
	\subsection{Overview}
\begin{figure}[t]
		\centering
		\setlength{\fboxrule}{0.pt}
		\setlength{\fboxsep}{0.pt}
		\fbox{
			\includegraphics[width=0.98\linewidth]{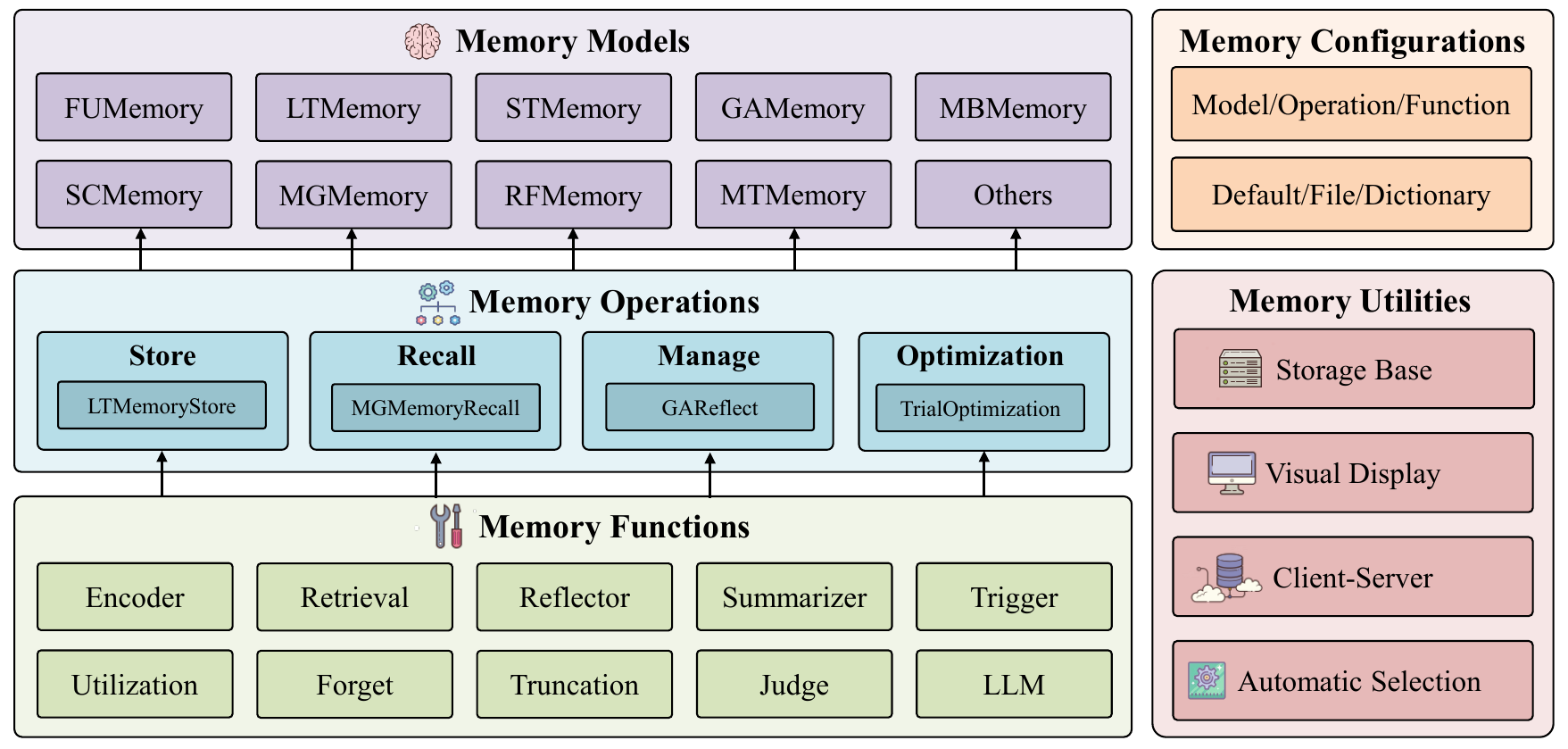}
		}
		\vspace{-0.65cm}
		\caption{An overview framework of MemEngine Library.}
		\vspace{-0.65cm}
		\label{fig:framework}
	\end{figure}

	The framework of our library is present in Figure~\ref{fig:framework}.
	The lowest level implements basic functions as standard support. The intermediate level comprises memory operations for fundamental processes. The highest level features various memory models from  previous research works. Additionally, we provide a configuration module and a utility module to facilitate convenient development and usage.
	
	\subsection{Memory Models}
	We implement a variety of memory models from recent research works under a general structure, allowing seamless switching among them. Specifically, these models are implemented with the interfaces including \textit{reset}, \textit{store}, \textit{recall}, \textit{manage}, and \textit{optimize}.
	
	The implemented memory models are described as follows:\\
	$\bullet$ \textbf{FUMemory} (Full Memory): Naively concatenates all the information into a single string, also known as long-context memory.\\
	$\bullet$ \textbf{LTMemory} (Long-term Memory): Calculates semantic similarities with text embeddings to retrieve the most relevant information. \\
	$\bullet$ \textbf{STMemory} (Short-term Memory): Maintains the most recent information and concatenates them into a single string.\\
	$\bullet$ \textbf{GAMemory} (Generative Agents~\cite{park2023generative}): A pioneer memory model with weighted retrieval combination and self-reflection mechanism.\\
	$\bullet$ \textbf{MBMemory} (MemoryBank~\cite{zhong2024memorybank}): A multi-layered memory model with dynamic summarization and forgetting mechanism.\\
	$\bullet$ \textbf{SCMemory} (SCM~\cite{wang2023enhancing}): A self-controlled memory model that can recall minimal but necessary information for inference.\\
	$\bullet$ \textbf{MGMemory} (MemGPT~\cite{packer2023memgpt}): A hierarchical memory model that treats the memory system as an operation system.\\
	$\bullet$ \textbf{RFMemory} (Reflexion~\cite{shinn2024reflexion}): A prominent memory model that can learn to memorize from previous trajectories by optimization.\\
	$\bullet$ \textbf{MTMemory} (MemTree~\cite{rezazadeh2024isolated}): A dynamic memory model with a tree-structured semantic representation to organize information.
	
	All of these memory models are implemented by combining various memory operations, and we make some reasonable adaptations in their implementations. Further details can be found in our project documentation and source code.
	
	\subsection{Memory Operations}
	We implement various types of memory operations for constructing memory models, including store, recall, manage, and optimize.
	
	\textbf{Memory Store Operation} intends to receive observations from the environment, processing them to obtain memory contents and adding them into memory storage. Another critical function of the memory store operation is to establish foundations for the memory recall operation, such as creating indexes and summaries.
	
	\textbf{Memory Recall Operation} intends to obtain useful information to assist agents in their decision-making. Typically, the input is a query or observation representing the current state of agents. Some human-like agents may also endow the memory recall operation with certain retention characteristics like human memory.
	
	\textbf{Memory Manage Operation} intends to reorganize existing information for better utilization, such as memory reflection. Besides, simulation-orientated agents may be equipped with a forgetting mechanism during the memory manage operation.
	
	\textbf{Memory Optimize Operation} intends to optimize the memory capability of LLM-based agents by using extra trials and trajectories. It enables agents to extract meta-insight from historical experiences, which can be considered as a learn-to-memorize procedure.
	
	Different memory models may share common memory operations or implement their unique operations  according to their requirements. For example, \textit{MTMemory} and \textit{LTMemory} share the common memory recall operation \textit{LTMemoryRecall}, while MTMemory has its own memory store operation \textit{MTMemoryStore} to implement the tree-structured information update.
	
	\subsection{Memory Functions}
	We implement various types of memory functions to support the construction of memory operations, which are listed as follows.
	
	\textbf{Encoder} can transfer textual messages into embeddings to represent in latent space by pre-trained models, such as E5~\cite{wang2022text}.
	
	\textbf{Retrieval} is utilized to find the most useful information for the current query or observation, commonly by different aspects like semantic relevance, importance, recency, and so on.
	
	\textbf{Reflector} aims to draw new insights in a higher level from existing information, commonly for reflection and optimization.
	
	\textbf{Summarizer} can summarize texts into a brief summary, which can decrease the lengths of texts and emphasize critical points.
	
	\textbf{Trigger} is designed to call functions or tools in extensible manners. One significant instance is utilizing LLMs to determine which function should be called with certain arguments. 
	
	\textbf{Utilization} aims to deal with several different parts of memory contents, formulating this information into a unified output.
	
	\textbf{Forget} is typically applied in simulation-oriented agents, such as role-playing and social simulations. It empowers agents with features of human cognitive psychology, aligning with human roles.
	
	\textbf{Truncation} helps to formulate memory contexts under the limitations of token numbers by certain LLMs.
	
	\textbf{Judge} intends to assess given observations or intermediate messages on certain aspects. For example, \textit{GAMemory} judges the importance score of each observation when stored into memory, as an auxiliary criterion for the retrieval process.
	
	\textbf{LLM} provides a convenient interface to utilize the powerful capability of different large language models.
	
	All these memory functions are designed to conveniently construct different memory operations for various models. For example, \textit{GAMemoryStore} utilizes \textit{LLMJudge} to provide the importance score on each observation.
	
	\subsection{Memory Configurations}
	In order to improve convenience for developers and facilitate parameter tuning by researchers, we have developed a unified memory configuration module. 
	First, we design a hierarchical memory configuration module corresponding to our three-level memory framework, enabling adjustments to both hyper-parameters and prompts within the memory models.
	Second, we provide a comprehensive set of default hyper-parameters and prompts, where developers and researchers can adjust only the specific parts without altering others.
	Finally, our configuration supports both statistic manners (\textit{e.g.,} files) and dynamic manners (\textit{e.g.,} dictionaries).
	
	\subsection{Memory Utilities}
	We implement extra utilities as auxiliary components, which are loosely coupled with the above modules. We implement a storage module as a database to retain the contents of information. We also provide a display module to visualize the specific contents within the memory. Besides, we offer a client module to utilize MemEngine through remote deployment on a server implemented by FastAPI.
	We also implement an automatic selector to assist developers to choose memory models with hyper-parameters for their own tasks.
	
	\section{Usage of MemEngine}
	In this section, we describe the usage of MemEngine to empower LLM-based agents with advanced memory capabilities. We divide our usage into two aspects: (1) utilize pre-implemented memory models, and (2) customize new memory models.
	
	\subsection{Utilize Pre-implemented Memory Models}
	There are two primary ways to deploy our library as follows.
	
	\textbf{Local Deployment.} Developers can easily install our library in their Python environment via pip, conda, or from source code. Then, they can create memory modules for their agents, and utilize unified interfaces to perform memory operations within programs.
	
	\textbf{Remote Deployment.} Alternatively, developers can install our library on computing servers and launch the service through a port. Then, they can initiate a client to perform memory operations by sending HTTP requests remotely from their lightweight devices.
	
	After deployment, there are three modes available for utilizing pre-implemented memory models. In the default mode, the library provides a comprehensive set of hyper-parameters and prompts for default usage. In the configurable mode, developers can adjust certain hyper-parameters and prompts to better adapt to their applications. In the automatic mode, the library automatically selects the appropriate memory models, hyper-parameters, and prompts from the provided ranges, based on a specific task's criteria.
	
	Additionally, our library offers compatibility with several well-known tools and frameworks, such as vllm and AutoGPT.
	
	\subsection{Customize New Memory Models}
	Our library provides support for developers to customize advanced memory models, offering comprehensive documentation and examples. There are three major aspects to customizing new models.
	
	\textbf{Customize Memory Functions.} Researchers may need to implement new functions in their models to extend existing ones for additional features. For example, they may extend \textit{LLMJudge} to design a \textit{BiasJudge} for poisoning detection.
	
	\textbf{Customize Memory Operations.} In developing a new model, customizing memory operations is crucial as they constitute the major pipelines of the detailed processes.
	For instance, a new memory recall operation can be implemented with a series of memory functions with advanced design and combination.
	
	\textbf{Customize Memory Models.}
	By integrating newly customized memory operations with existing ones, researchers can design their models with various combinations to best suit their applications.

	\section{Conclusion}
	In this paper, we introduce a unified and modular library for developing advanced memory of LLM-based agents. In the future, we plan to provide support for multi-modal memory (such as visual and audio memory) to further enrich and enhance the memory capabilities of LLM-based agents for wider applications.
	
	\begin{acks}
		This work is supported in part by National Natural Science Foundation of China (No. 62422215 and No. 62472427), Beijing Outstanding Young Scientist Program NO.BJJWZYJH012019100020098, Intelligent Social Governance Platform, Major Innovation \& Planning Interdisciplinary Platform for the “DoubleFirst Class” Initiative, Renmin University of China, Public Computing Cloud, Renmin University of China, fund for building world-class universities (disciplines) of Renmin University of China, Intelligent Social Governance Platform.
		This work is also sponsored in part by Huawei Innovation Research Program. We gratefully acknowledge the support from Mindspore\footnote{\url{https://www.mindspore.cn/}}, CANN (Compute Architecture for Neural Networks) and Ascend AI Processor used for this research.
	\end{acks}
	
	\bibliographystyle{ACM-Reference-Format}
	\balance
	\bibliography{reference.bib}

\end{document}